\documentclass{article}
\usepackage{spconf}
\usepackage[bookmarks=false,hyperfigures=false,hidelinks=true]{hyperref}
\usepackage{amsmath}
\usepackage{amssymb, subfigure, url, doi, xspace}
\usepackage{mathtools}
\usepackage{sidecap}
\ninept

\sidecaptionvpos{figure}{t}

\usepackage{algorithm2e}
\usepackage{listings}
\usepackage{ifpdf}

\ifpdf
  \usepackage[final]{graphicx}
  \DeclareGraphicsExtensions{.pdf,.png}
\else
  \usepackage[final]{graphicx}
  \DeclareGraphicsExtensions{.pstex,.ps,.eps,.png}
\fi
\usepackage{color}

\newcommand{\ie}{i.e.\xspace}

\newcommand{\eq}[1]{(\ref{eq:#1})}
\newcommand{\fig}[1]{Fig.~\ref{fig:#1}}
\newcommand{\figs}[1]{Figs.~\ref{fig:#1}}
\newcommand{\fign}[1]{\ref{fig:#1}}
\newcommand{\tbl}[1]{Table~\ref{tbl:#1}}

\newcommand{\smallmath}[1]{%
  {\small
    \setlength{\abovedisplayskip}{6pt}
    \setlength{\belowdisplayskip}{\abovedisplayskip}
    \setlength{\abovedisplayshortskip}{0pt}
    \setlength{\belowdisplayshortskip}{3pt}
    #1
  }%
}%

\makeatletter%
\if@twocolumn%
\else%
\fi
\makeatother

\allowdisplaybreaks[4]

\newcommand{\mb}[1]{\mathbf{#1}}
\newcommand{\mbs}[1]{\boldsymbol{#1}}

\newcommand{\mc}[1]{\mathcal{#1}}

\newcommand{\sign}{\mathop{\mathrm{sign}}}

\newcommand{\norm}[1]{\left\| #1 \right\|}
\DeclarePairedDelimiterX{\normsz}[1]{\lVert}{\rVert}{#1}
\newcommand{\abs}[1]{\left| #1 \right|}

\DeclareMathOperator*{\argmin}{arg\,min}

\date{\today}

\begin{document}

\title{Convolutional Sparse Coding: Boundary Handling Revisited}

\twoauthors
{Brendt Wohlberg%
  \sthanks{This research was supported by the
    U.S. Department of Energy via the LANL/LDRD Program.}}
{Theoretical Division\\
  Los Alamos National Laboratory\\
  Los Alamos, NM 87545, USA}
{Paul Rodr\'{i}guez%
\sthanks{This research was supported by the ``Programa Nacional de Innovaci\'on
para la Competitividad y Productividad''
(Inn\'ovate Per\'u) Program.}}
{Department of Electrical Engineering\\
  Pontificia Universidad Cat\'olica del Per\'u\\
  Lima, Peru}

\maketitle

\begin{abstract}
  Two different approaches have recently been proposed for boundary handling in convolutional sparse representations, avoiding potential boundary artifacts arising from the circular boundary conditions implied by the use of frequency domain solution methods by introducing a spatial mask into the convolutional sparse coding problem. In the present paper we show that, under certain circumstances, these methods fail in their design goal of avoiding boundary artifacts. The reasons for this failure are discussed, a solution is proposed, and the practical implications are illustrated in an image deblurring problem.
\end{abstract}


\section{Introduction}
\label{sec:intro}

Convolutional sparse representations~\cite{lewicki-1999-coding, zeiler-2010-deconvolutional} provide a convenient alternative to the standard approach of block-based application of sparse representations to images.  With the recent progress in the development of efficient algorithms for convolutional sparse coding (CSC)~\cite{chalasani-2013-fast, bristow-2013-fast, wohlberg-2014-efficient, wohlberg-2016-efficient}, this form of sparse representations has become a practical approach for imaging inverse problems. A critical component of the current state of the art algorithms for CSC is the handling of the convolution in the frequency domain, which automatically implies %
circular boundary conditions. Recently, two different approaches have been proposed for avoiding boundary artifacts by performing a boundary extension and solving the CSC problem with a mask on the critical region in the data fidelity term. One of these proposes application of the mask decoupling (MD) technique~\cite{almedia-2013-deconvolving} to the CSC problem~\cite{heide-2015-fast}.  The other, additive mask simulation (AMS)~\cite{wohlberg-2016-boundary} takes advantage of the particular form of the CSC problem to represent the multiplicative mask by introducing an additional dictionary filter which is constrained in a way that simulates that mask.

\section{Convolutional Sparse Coding}
\label{sec:csc}

By far the most common form of CSC is
\vspace{-1.2mm}
\begin{equation}
\argmin_{\{\mb{x}_m\}} \frac{1}{2} \normsz[\Big]{\sum_m \mb{d}_m \ast \mb{x}_m
- \mb{s}}_2^2 + \lambda \sum_m \alpha_m \norm{\mb{x}_m}_1 \; ,
\label{eq:convbpdn}
\vspace{-1mm}
\end{equation}
where $\{\mb{d}_m\}$ is a set of $M$ dictionary \emph{filters}, $\ast$ denotes convolution, $\{\mb{x}_m\}$ is a set of coefficient maps, and the $\alpha_m$ allow distinct weighting of the $\ell_1$ term for each filter $\mb{d}_m$. Defining $D_m$ is a linear operator such that $D_m \mb{x}_m = \mb{d}_m \ast \mb{x}_m$, and defining block matrices and vectors \vspace{-2mm} \smallmath{ \addtolength{\arraycolsep}{-1mm}
\begin{equation}
D = \left( \begin{array}{ccc}D_0 & D_1 &
    \ldots \end{array} \right)  \;\;
\mbs{\alpha} = \left( \begin{array}{c}  \alpha_0 I\\ \alpha_1 I\\
    \vdots  \end{array} \right)
\;\;
\mb{x} = \left( \begin{array}{c}  \mb{x}_0\\ \mb{x}_1\\
    \vdots  \end{array} \right)
\vspace{-0.2mm}
\label{eq:dxcbpdn}
\end{equation}
}%
allows~\eq{convbpdn} to be written in the form
\vspace{-2mm}
\begin{equation}
\argmin_{\mb{x}} \frac{1}{2} \norm{D \mb{x} - \mb{s}}_2^2 + \lambda
\norm{\mbs{\alpha} \odot \mb{x}}_1 \;,
\label{eq:bpdn}
\vspace{-1mm}
\end{equation}
where $\odot$ denotes the Hadamard product.  This problem can be solved within the Alternating Direction Method of Multipliers (ADMM)~\cite{boyd-2010-distributed} framework by an iterative scheme for the equivalent constrained problem \vspace{-2.6mm}
\begin{equation}
\argmin_{\mb{x},\mb{y}} \frac{1}{2} \normsz[\big]{
    D \mb{x} - \mb{s}}_2^2 \!+\! \lambda
  \norm{\mbs{\alpha} \odot \mb{y}}_1 \text{ s.t. } \mb{x} \! = \!
  \mb{y}  \; .
\label{eq:cbpdnsplit}
\vspace{-1mm}
\end{equation}
The iterative scheme solves two subproblems, one associated with the data fidelity term, which is solved in the frequency domain~\cite{wohlberg-2016-efficient}, and the other involving the regularization term, which is solved via the closed-form expression for soft-thresholding.

Boundary masking involves introducing a diagonal matrix, $W$, implementing a spatial mask, into the data fidelity term, as in \vspace{-1.2mm}
\begin{equation}
\argmin_{\mb{x}} \frac{1}{2} \normsz[\big]{W D \mb{x}
- \mb{s}}_2^2 + \lambda \norm{\mbs{\alpha} \odot \mb{x}}_1 \; .
\label{eq:convbpdnws}
\vspace{-0.3mm}
\end{equation}
The presence of the spatial mask prevents direct application of the frequency domain solution since, unlike the block components $D_m$ of $D$, $W$ is not diagonal in the frequency domain.  The MD approach avoids this difficulty by replacing~\eq{bpdn} with a different constrained problem~\cite{heide-2015-fast} \vspace{-1.2mm}
{\small \setlength{\abovedisplayskip}{6pt} \setlength{\belowdisplayskip}{\abovedisplayskip} \setlength{\abovedisplayshortskip}{0pt} \setlength{\belowdisplayshortskip}{3pt}
\begin{equation}
\hspace{-2mm}%
\begin{multlined}
\argmin_{\mb{x},\mb{y}_0,\mb{y}_1} \frac{1}{2}
\normsz[\big]{W \mb{y}_1 - \mb{s}}_2^2  \!+\! \lambda
  \norm{\mbs{\alpha} \odot \mb{y}_0}_1 \\
  \text{ s.t. }
  \left( \begin{array}{c} \mb{x} \\
    D \mb{x} \end{array} \right)
  -
 \left( \begin{array}{c} \mb{y}_0 \\ \mb{y}_1 \end{array} \right)
 = 0
  \; .
\end{multlined}
\label{eq:cbpdnsplitmd}
\end{equation}
}%
This form decouples $W$ from the sum of convolutions $D \mb{x}$, facilitating the use of the same frequency domain solution used for~\eq{cbpdnsplit}.

The iterations of the ADMM algorithm for problem~\eq{cbpdnsplitmd} (see~\cite[Sec. 4.2]{wohlberg-2016-boundary}) involve the following updates \vspace{-1mm}
\begin{align}
(D^T D + I) \mb{x}^{(j+1)} &= D^T \big(\mb{y}_1^{(j)} - \mb{u}_1^{(j)}\big) +
\big(\mb{y}_0^{(j)} - \mb{u}_0^{(j)}\big) \label{eq:mdx} \\
\mb{y}_0^{(j+1)} &= \; \mc{S}_{\lambda \mbs{\alpha} / \rho}\big( \mb{x}^{(j+1)} +
   \mb{u}_0^{(j)} \big) \label{eq:mdy0} \\
(W^T W + \rho I) \mb{y}_1^{(j+1)} &= W^T \mb{s} + \rho \big(D
\mb{x}^{(j+1)}  + \mb{u}_1^{(j)}\big) \label{eq:mdy1} \\
\mb{u}_0^{(j+1)} &= \mb{u}_0^{(j)} + \mb{x}^{(j+1)} - \mb{y}_0^{(j+1)}
\label{eq:mdu0}\\
\mb{u}_1^{(j+1)} &= \mb{u}_1^{(j)} + D \mb{x}^{(j+1)} -
\mb{y}_1^{(j+1)} \label{eq:mdu1}
\;,
\end{align}
where the iteration index is indicated by a superscript in parentheses and
\vspace{-1mm}
\begin{equation}
\mc{S}_{\mbs{\gamma}}(\mb{u}) = \sign(\mb{u}) \odot \max(0, \abs{\mb{u}}
  - \mbs{\gamma}) \;.
\label{eq:shrink}
\end{equation}
Update~\eq{mdx} can be solved in the frequency domain as in~\cite{wohlberg-2014-efficient},~\eq{mdy0} is solved via the closed-form expression for soft-thresholding, and~\eq{mdy1} is a computationally cheap linear problem since $W$ is diagonal.

The AMS method~\cite{wohlberg-2016-boundary} takes a fundamentally different approach, retaining the original constrained form~\eq{cbpdnsplit}, but introducing into the representation an additive component that is constrained to be zero within the active part of the mask, and is unconstrained and un-penalized within the masked-out region. Due to length restrictions, this method is not discussed in further detail here, and all results presented in the following sections are computed using the MD method. It should be noted, though, that the AMS method is prone to the same issues, and amenable to similar solutions.

\section{Boundary Masking Failure}
\label{sec:mskfail}

In principle these masking techniques provide for a complete decoupling between the active and masked-out regions of the solution, but, in practice, the effects of the presence of a boundary can propagate into the active region of the solution under certain circumstances.

\subsection{Phenomenon}

\begin{SCfigure}[50][htpb]
  \centering \small
  \vspace{-9mm}
  \begin{minipage}[t]{3cm}
  \vspace{3mm}
  \includegraphics[width=3cm]{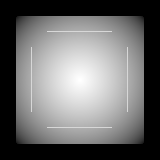}
  \end{minipage}
  \caption{Synthetic $160 \times 160$ pixel test image constructed from a $128 \times 128$ pixel image consisting of a smooth Gaussian surface together with two horizontal and two vertical edges, created by setting the corresponding pixels to unit value. The final image is obtained by zero-padding by 16 pixels on all sides.}
  \label{fig:syntst}
\end{SCfigure}

This effective failure of boundary masking is most easily illustrated with a simple synthetic test case\footnote{This is not a contrived example: it was constructed with the goal of understanding a phenomenon that was first observed while attempting to solve a practical inverse problem involving natural images.}. The test image was as in~\fig{syntst}, and a corresponding mask matrix, $W$, was defined to represent a mask that is zero in the padded region and unity elsewhere.  The dictionary consisted of three $16 \times 16$ sample filters, a Gaussian surface, $\mb{d}_0$, and horizontal and vertical lines, $\mb{d}_1$ and $\mb{d}_2$ respectively.  The corresponding weights were set as $\alpha_0 = 0, \alpha_1 = \alpha_2 = 1$, \ie no regularisation on the coefficient map for the Gaussian filter since its role was representing the smooth component of the image. This setup provides a simplified cartoon of convolutional decomposition of a natural image using a dictionary consisting of a single smooth filter for representing the smooth component of the image and a potentially large number of learned filters for representing the image texture and edges~\cite[Sec. 3]{wohlberg-2016-convolutional2}.

\begin{figure}[htbp]
  \centering \small
  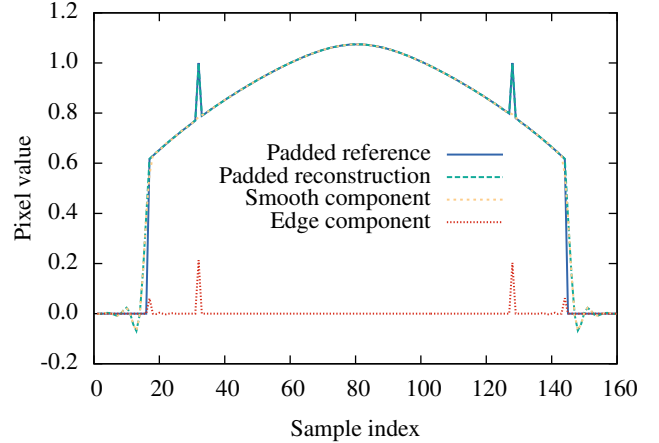
  \caption{Cross section of the test image, its reconstruction from the sparse representation, and the distinct contributions to the reconstruction from the smooth and edge filters, \ie $\mb{d}_0 \ast \mb{x}_0$ and $\sum_{m=1}^2 \mb{d}_m \ast \mb{x}_m$ respectively. Sparse representation computed using 500 iterations of the MD algorithm with $\mb{y}_1$ initialised as the zero padded input signal.}
  \label{fig:mskfail}
\end{figure}

\begin{figure}[htbp]
  \centering \small
  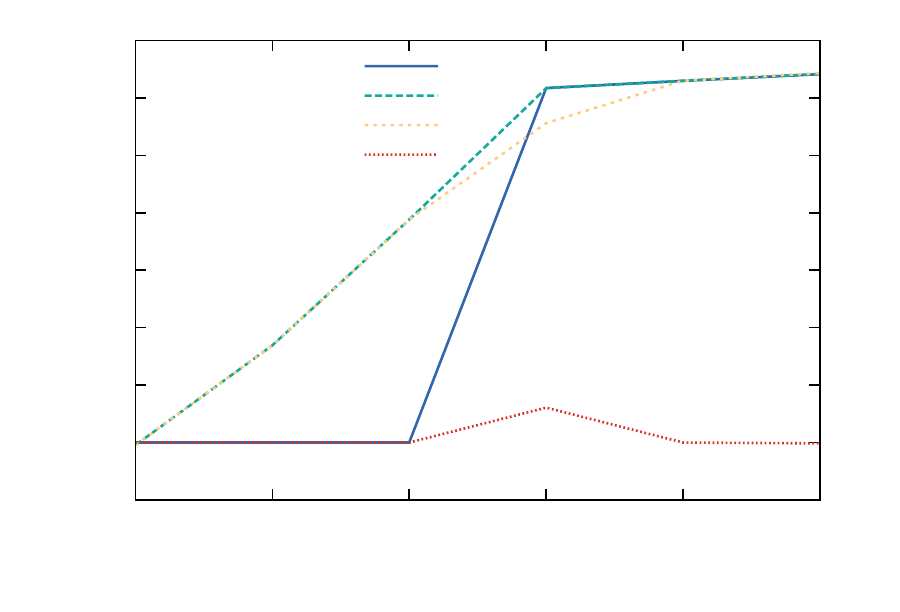
  \caption{A zoom into the left boundary region of~\fig{mskfail}.}
  \label{fig:mskfailcrop}
\end{figure}

Problem~\eq{cbpdnsplitmd} was solved via 500 iterations of the ADMM algorithm for the MD approach\footnote{Computed using the Matlab version of SPORCO library~\cite{wohlberg-2016-sporco}.}. In addition to the full signal reconstruction, $\sum_m \mb{d}_m \ast \mb{x}_m$, the individual contribution to this reconstruction from each filter, $\mb{d}_m \ast \mb{x}_m$, was also computed.  A cross section of these results is displayed in~\fig{mskfail}, and a zoom into the left boundary region is displayed in~\fig{mskfailcrop}.  The boundary masking failure is subtle in that it is not visible if we only look at the full signal reconstruction, $\sum_m \mb{d}_m \ast \mb{x}_m$, since the reconstruction matches the reference to very high accuracy all the way to the boundary of the masked region. The boundary artifacts only become visible if we examine the individual components, $\mb{d}_m \ast \mb{x}_m$: here we see that the smooth component starts to decay \emph{before} the boundary is reached, with the accuracy of the final reconstruction being maintained by a compensating peak in the edge component, also \emph{within} the active region of the mask. Since the overall sum is accurate, these artifacts in the individual components do not matter if the image is directly reconstructed from its sparse representation, but the sparse representation is usually computed as part of the solution of some inverse problem. In this scenario the coefficient maps are processed prior to reconstruction, so that canceling of the component artifacts to give an accurate reconstruction is no longer guaranteed, risking artifacts in the final output image.

\subsection{Initialisation}

\begin{figure}[htbp]
  \centering \small
  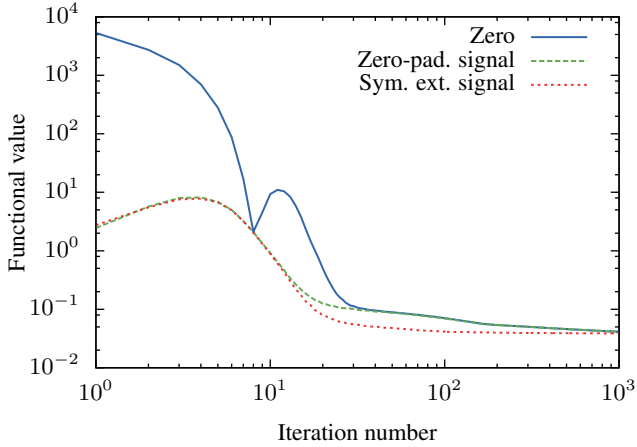
  \caption{A comparison of functional convergence of the MD method with three different initialisations for $\mb{y}_1$: a zero vector, a zero padded version of the input signal, and a symmetrically extended version of the input signal.}
  \label{fig:fnccnv}
\end{figure}

\begin{figure}[htbp]
  \centering \small
  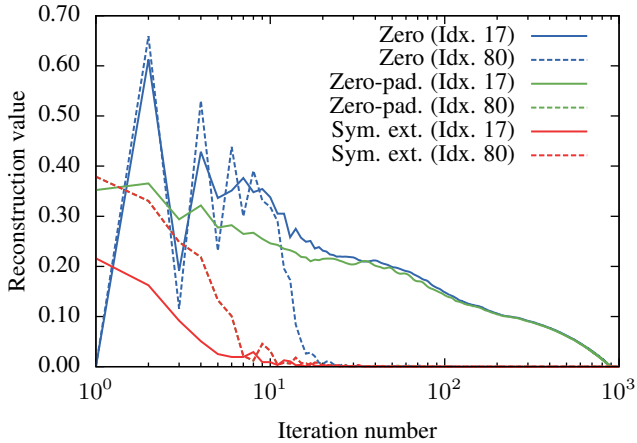
  \caption{A comparison of the evolution of two different locations in the reconstructed edge component for the same initialisation choices considered in~\fig{fnccnv}. The locations are at sample indices 17 (artifact location) and 80 (signal centre). The curve for ``Zero-pad. (Idx. 80)'' is not visible as it coincides with that for ``Sym. ext. (Idx. 80)''.}
  \label{fig:smpcnv}
\end{figure}

This phenomenon is rather surprising since the functional we minimize is specifically designed to avoid boundary artifacts.  To understand and address it, we need to consider the choice of initial values for variables $\mb{y}_0, \mb{y}_1, \mb{u}_0$, and $\mb{u}_1$ in the MD algorithm. This choice is not discussed at all in~\cite{heide-2015-fast}, but it is a reasonable assumption (supported by inspection of the corresponding publicly available software) that the authors follow the common approach for ADMM algorithms in setting them all to zero vectors. Although not discussed in~\cite{wohlberg-2016-boundary} either, the actual experiments reported in that work (see~\cite{wohlberg-2016-sporco}) made use of an alternative initialisation, setting $\mb{y}_1$ to the input signal $\mb{s}$. A third initialisation strategy is proposed here: setting $\mb{y}_1$ to a version of the signal that has been extended symmetric extension instead of zero-padding.

\begin{figure}[htbp]
  \centering \small
  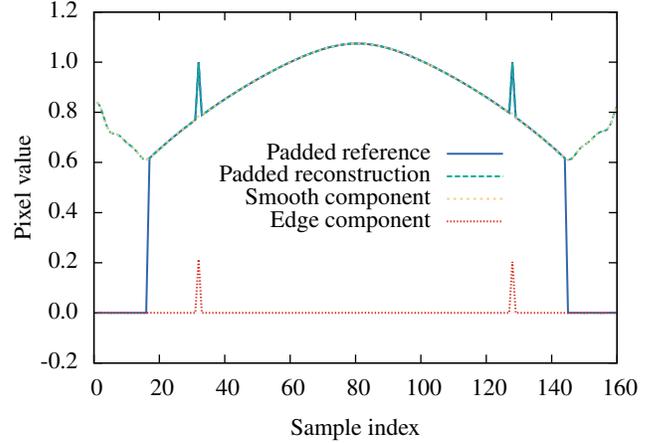
  \caption{Cross section of the test image, its reconstruction from the sparse representation, and the distinct contributions to the reconstruction from the smooth and edge filters, \ie $\mb{d}_0 \ast \mb{x}_0$ and $\sum_{m=1}^2 \mb{d}_m \ast \mb{x}_m$ respectively. Sparse representation computed using 100 iterations of the MD algorithm with $\mb{y}_1$ initialised as the symmetrically extended input signal. Note the residual effect of the $\mb{y}_1$ initialisation on the reconstruction in the masked-out region.}
  \label{fig:mskfix}
\end{figure}

\begin{figure}[htbp]
  \centering \small
  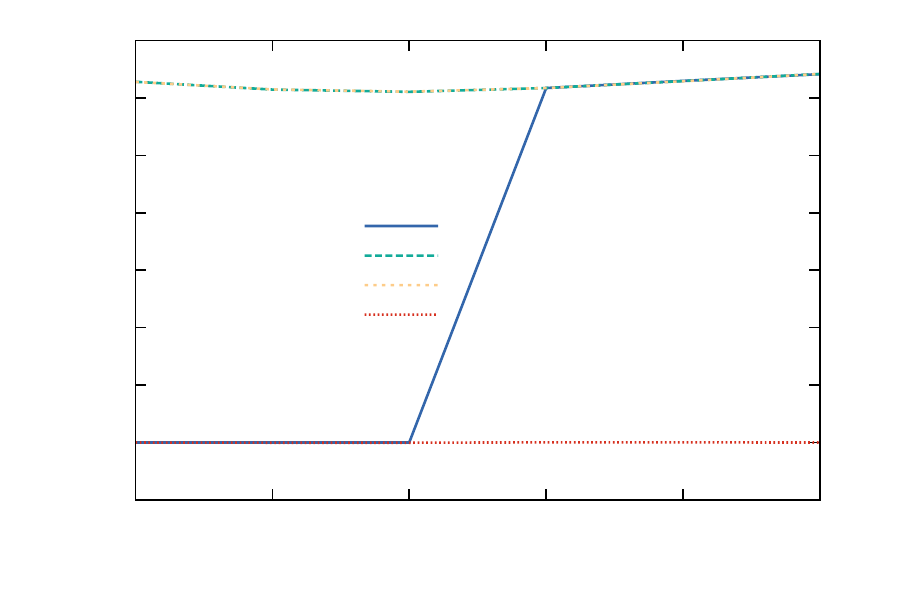
  \caption{A zoom into the left boundary region of~\fig{mskfix}.}
  \label{fig:mskfixcrop}
\end{figure}

The convergence of the MD algorithm with these three initialization choices, for the synthetic test problem, is compared in~\fig{fnccnv}. It can be seen that the two signal-based initialisations are initially quite similar, and substantially better than zero initialisation, but the symmetric extension exhibits better behaviour after 100 iterations. The corresponding evolution of the reconstructed edge component, $\sum_{m=1}^2 \mb{d}_m \ast \mb{x}_m$, at the location of the artifact (see~\figs{mskfix} and~\fign{mskfixcrop}) and at the signal centre are compared in~\fig{smpcnv}. The cause of the apparent boundary masking failure is now clear: the reconstructed edge component converges very slowly to its ``correct'' value for both the zero and the zero padded initialisations. The symmetric-extension initialisation, in contrast, results in convergence at the boundary that is comparable to that at the signal centre, far removed from the boundary effects.  With the use of this initialisation, the boundary artifact is no longer apparent in~\fig{mskfix} and~\fig{mskfixcrop} after only 100 iterations.

The effect of the initial choice of $\mb{y}_1$ can be understood by referring to the optimisation problem~\eq{convbpdnws} and the corresponding ADMM iterations~\eq{mdx}--\eq{mdu1}. The coefficient map, $\mb{x}_0$, of the smooth dictionary filter, $\mb{d}_0$, has zero weight in the regularization term ($\alpha_0 = 0$) and its reconstruction, $\mb{d}_0 * \mb{x}_0$, has no influence on the data fidelity term outside of the active region of the mask matrix, $W$. The solution for the part of this coefficient map for which $\mb{d}_0 * \mb{x}_0$ lies outside of the active mask region is therefore completely undetermined, all possible choices having equal cost. It can be seen from~\eq{mdx} that the initial choice of $\mb{y}_1$ determines the initial solution $\mb{x}$ for the entire domain, \ie with no application of the mask matrix $W$. The part of this initial solution that lies outside of the active region of the mask persists for the remainder of the iterations since~\eq{mdy1} does not modify $\mb{y}_1$ outside of the active region. As a result, the solution to which ADMM algorithm~\eq{mdx}--\eq{mdu1} converges depends on the choice of the initial value for $\mb{y}_1$, as can be seen by comparing the ``smooth component'' reconstructions in~\figs{mskfail} and~\fign{mskfix}.

\section{Gaussian Blur Deconvolution}
\label{sec:rslt}

The impact of the slow boundary convergence is illustrated in a Gaussian blur deconvolution problem. The $512 \times 512$ pixel greyscale ``Lena'' image with pixels rescaled to the range $[0,1]$ was used as a reference image $\mb{s}$. A test image, $\mb{s}_{\mathrm{bn}}$, was constructed by convolving $\mb{s}$ with a $7 \times 7$ sample Gaussian filter $\mb{h}$ with unit standard deviation parameter, adding Gaussian white noise with standard deviation 0.01, and zero-padding by 39 pixels on all sides. The corresponding mask matrix, $W$, was defined to represent a spatial mask set to be zero in the padded region and unity elsewhere.

The CSC-based deconvolution method involved solving problem~\eq{cbpdnsplitmd} with the test image, $\mb{s}_{\mathrm{bn}}$, and a blurred dictionary, $\{\mb{g}_m\}$, and constructing the estimate of the deblurred image as the reconstruction of the resulting coefficient maps with respect to a corresponding unblurred dictionary, $\{\mb{d}_m\}$, \ie
\begin{align}
\mb{x}_{\mathrm{b}} &= \argmin_{\mb{x}} \frac{1}{2} \normsz[\big]{W G \mb{x}
- \mb{s}_{\mathrm{bn}}}_2^2 + \lambda \norm{\mbs{\alpha} \odot
\mb{x}}_1 \\
\hat{\mb{s}} &= D \mb{x}_{\mathrm{b}} \;,
\end{align}
where $\hat{\mb{s}}$ is the deconvolved estimate of $\mb{s}$, and $D$ and $G$ are the block-matrix forms, as introduced in~\eq{dxcbpdn}, of dictionaries $\{\mb{d}_m\}$ and $\{\mb{g}_m\}$ respectively.  The reconstruction dictionary, $\{\mb{d}_m\}$, consisted of a smooth Gaussian filter of $64 \times 64$ samples (with standard deviation parameter 5.0) and a learned multiscale dictionary with 16 filters of $8 \times 8$ samples, 32 filters of $12 \times 12$ samples, and 48 filters of $16 \times 16$ samples. (As in the previous section, the component of $\mbs{\alpha}$ corresponding to the Gaussian dictionary component was set to zero.) The corresponding blurred dictionary $\{\mb{g}_m\}$ was obtained by convolving each filter in $\{\mb{d}_m\}$ by the blurring kernel $\mb{h}$, \ie $\mb{g}_m = \mb{h} \ast \mb{d}_m$, with appropriate zero padding to avoid boundary effects on the blurred dictionary filters.

\begin{figure}[htbp]
  \centering \small
  \begin{tabular}{cc}
    \subfigure[\label{fig:dcnvref}\protect\rule{0pt}{1.5em}
    Reference]
    {\includegraphics[width=4cm,trim=350 0 0 350,clip]{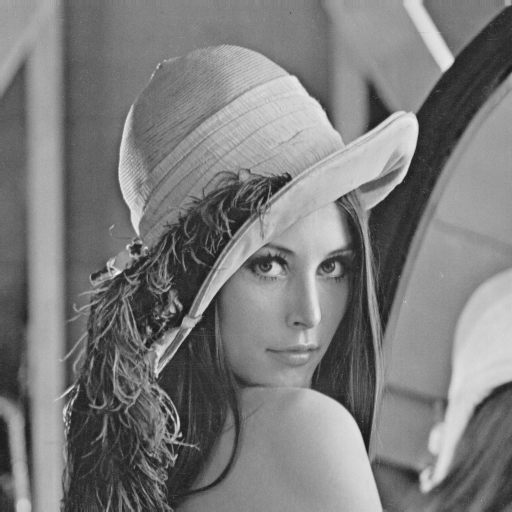}} &
    \subfigure[\label{fig:dcnvblr}\protect\rule{0pt}{1.5em}
    Blurred with noise]
    {\includegraphics[width=4cm,trim=350 0 0 350,clip]{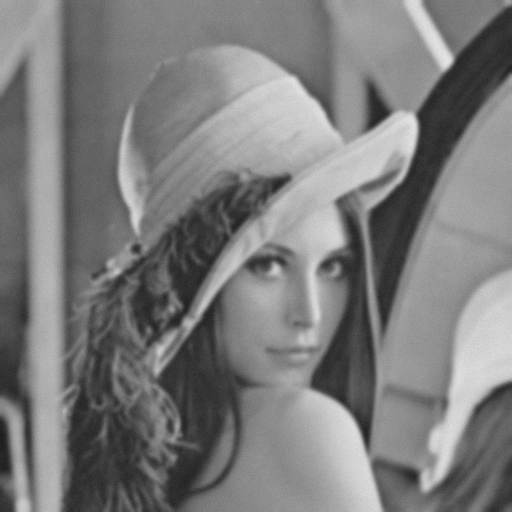}} \\
    \subfigure[\label{fig:cscbflcrp}\protect\rule{0pt}{1.5em}
    CSC-zp]
    {\includegraphics[width=4cm,trim=350 0 0 350,clip]{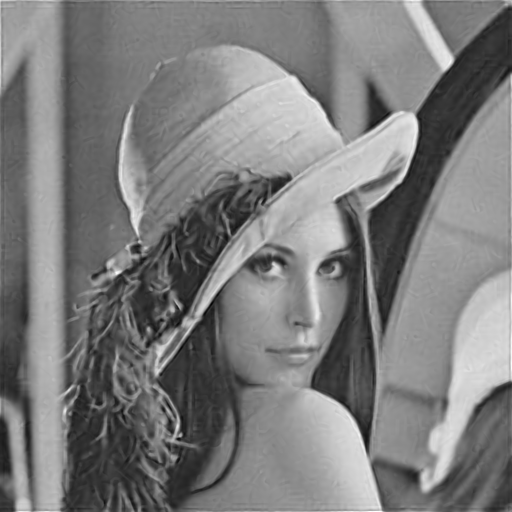}} &
    \subfigure[\label{fig:cscbfxcrp}\protect\rule{0pt}{1.5em}
    CSC-se]
    {\includegraphics[width=4cm,trim=350 0 0 350,clip]{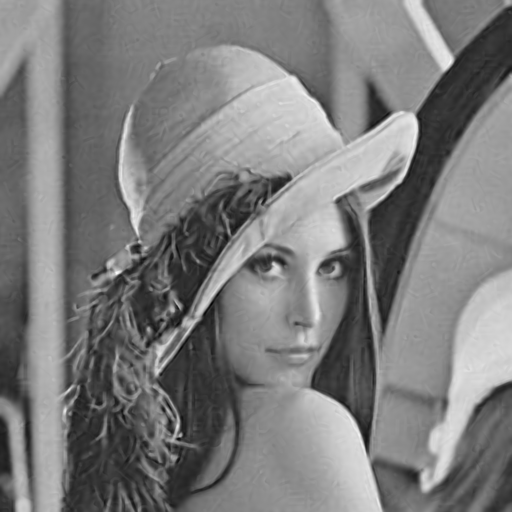}}
  \end{tabular}
  \vspace{-3mm}
  \caption{Bottom right corner of the deconvolution test images and reconstructions using two different initialisations (CSC-zp and CSC-se) for the MD algorithm.}
  \label{fig:dcnvcrp}
\end{figure}

Two different deconvolved estimates were computed, one using 500 iterations of the MD algorithm with $\mb{y}_1$ initialised using the zero padded test image (CSC-zp), and the other initialised using a version of the test image extended via symmetric extension (CSC-se).  The corresponding reconstruction PSNR values are presented in~\tbl{deconv}, and the bottom right corners of the two images are displayed in~\fig{dcnvcrp}. Note the very clear boundary artifacts in~\fig{cscbflcrp}, and the substantial effect on the overall reconstruction PSNR in~\tbl{deconv}, despite the occurrence of the boundary artifacts in a relatively small fraction of the entire image.

\begin{table}[htbp]
  \centering
  \begin{tabular}{|l|r|r|r|r|r|} \hline
   & \multicolumn{1}{|c|}{Test} &
     \multicolumn{1}{|c|}{CSC-zp} &
     \multicolumn{1}{|c|}{CSC-se} &
     \multicolumn{1}{|c|}{TV} &
     \multicolumn{1}{|c|}{EPLL} \\ \hline
  PSNR (dB) &  26.81 & 29.90 & 30.41 & 30.17 & 30.60 \\ \hline
 \end{tabular} \vspace{1mm}
  \caption{
   A comparison of PSNR values for the blurred and noisy test image
   and the deconvolved estimates obtained via the CSC MD algorithm
   with  two different initialisations (CSC-zp and CSC-se) as well as
   via the TV~\cite{dias-2006-total} and
   EPLL~\cite{zoran-2011-learning} methods.
  }
  \label{tbl:deconv}
\end{table}

Although the primary focus of these experiments is to demonstrate the impact of the boundary phenomenon discussed above, comparisons with the Total Variation (TV)~\cite{dias-2006-total} and Expected Patch Log Likelihood (EPLL)~\cite{zoran-2011-learning} methods are included in~\tbl{deconv} as a performance reference. CSC-zp has the worst PSNR, while that of CSC-se is intermediate between those of TV and EPLL.

\section{Conclusions}
\label{sec:cnclsn}

The mathematical formulation of boundary handling via a masked data fidelity term might lead one to conclude that the type of boundary extension is irrelevant to the solution since it lies outside of the masked region. The phenomenon of very slow convergence at the boundary when zero padding, however, contradicts this intuition and demonstrates that some care must be exercised in choosing the extension method. For the cases demonstrated here, initialisation of the MD algorithm via symmetric extension of the input signal completely suppresses the boundary artifacts that are observed when initialising to a zero vector or via zero padding of the input signal

\bibliographystyle{IEEEtranD}
\bibliography{paper}

\end{document}